\documentclass[conference]{IEEEtran}
\IEEEoverridecommandlockouts
\usepackage{cite}
\usepackage{amsmath,amssymb,amsfonts}
\usepackage{algorithmic}
\usepackage{graphicx}
\usepackage{textcomp}
\usepackage{xcolor}
\def\BibTeX{{\rm B\kern-.05em{\sc i\kern-.025em b}\kern-.08em
		T\kern-.1667em\lower.7ex\hbox{E}\kern-.125emX}}
\begin{document}
	
	\title{On the Vietnamese Name Entity Recognition: A Deep Learning Method Approach}

	\author{\IEEEauthorblockN{Ngoc C. L\^{e}}
		\IEEEauthorblockA{\textit{School of Applied Mathematics and Informatics} \\
			\textit{Hanoi University of Science and Technology} \\
			\textit{Institution of Mathematics} \\
			\textit{Vietnam Academy of Science and Technology} \\
			lechingoc@yahoo.com}
		\and
		\IEEEauthorblockN{Ngoc-Yen Nguyen}
		\IEEEauthorblockA{\textit{School of Applied Mathematics and Informatics}\\
			\textit{Hanoi Univ of Science and Technology} \\
			yen.nn154447@sis.hust.edu.vn}
		\and 
		\IEEEauthorblockN{Anh-Duong Trinh}
		\IEEEauthorblockA{\textit{iCOMM Media \(\& \) Tech, Jsc}\\
			duong.trinh@icomm.vn}
		
	}
	\maketitle
	\begin{abstract}
		Named entity recognition (NER) plays an important role in text-based information retrieval. In this paper, we combine Bidirectional Long Short-Term Memory (Bi-LSTM) \cite{hochreiter1997,schuster1997} with Conditional Random Field (CRF) \cite{lafferty2001} to create a novel deep learning model for the NER problem. Each word as input of the deep learning model is represented by a Word2vec-trained vector. A word embedding set trained from about one million articles in 2018 collected through a Vietnamese news portal (baomoi.com). In addition, we concatenate a Word2Vec\cite{mikolov2013}-trained vector with semantic feature vector (Part-Of-Speech (POS) tagging, chunk-tag) and hidden syntactic feature vector (extracted by Bi-LSTM nerwork) to achieve the (so far best) result in Vietnamese NER system. The result was conducted on the data set VLSP2016 (Vietnamese Language and Speech Processing 2016 \cite{vlsp2016}) competition.
	\end{abstract}
	
	\begin{IEEEkeywords}
		Vietnamese, Named Entity Recognition, Long Short-Term Memory, Conditional Random Field, Word Embedding
	\end{IEEEkeywords}
	
	\section{Introduction}
	Named-entity recognition (NER) (also known as entity identification, entity chunking and entity extraction) is a subtask of information extraction that seeks to locate and classify named entity mentions in unstructured text into pre-defined categories such as the person names, organizations, locations, medical codes, time expressions, quantities, monetary values, percentages, etc . It is a fundamental NLP research problem that has been studied for years. It is also considered as one of the most basic and important tasks in some big problems such as information extraction, question answering, entity linking, or machine translation. Recently, there are many novel ideal in NER task such as Cross-View Training (CVT) \cite{clark2018}, a semi-supervised learning algorithm that improves the representations of a Bi-LSTM sentence encoder using a mix of labeled and unlabeled data, or deep contextualized word representation \cite{peters2018} and contextual string embeddings, a recent type of contextualized word embedding that were shown to yield state-of-the-art results \cite{akbik2018,akbik2019}. These studies have shown new state-of-the-art methods with F1 scores on NER task.\\
	In Vietnamese language, NER systems in VLSP 2016 adopted either conventional feature-based sequence labeling models such as Recurrent neural network (RNN), Bidirectional Long Short Term Memory (Bi-LSTM) \cite{pham2017}, Maximum-Entropy-Markov Models (MEMMs) \cite{lehong2016,nguyen2016}, Conditional Random Fields (CRFs) \cite{le2016,nguyen2016-2}. For the VLSP 2016 data set, the first Vietnamese NER system has applied MEMMs with specific features \cite{pham2017}. However, they have not achieved accuracy that far beyond those of classical machine learning methods. Most of the above models depends heavily on specific resources and hand-crafted features, which makes it difficult for those models to apply to new domains and other tasks.\\
	In \cite{minh2018,minh2018-2}, the author used the information of word, word shapes, part-of-speech tags, chunking tags as hand-crafted features for CRF to label entity tags \cite{nguyen2019}. Over the past few years, many deep learning models have been proposed to overcome these limitations. Some NER models have used LSTM and CRF to predict NER \cite{huang2015,le2017}. In addition, benefits from both the expression of words and characters when combining CNN and CRF are presented in \cite{ma2016,wu2019}.\\
	In this study, we introduce a deep neural network for Vietnamese NER using extraction of morphological features automatically through a Bi-LSTM (character feature) network combined with POS features - tagging and chunk tag. The model includes two bidirectional-lstm hidden layer and an output layer CRF. For Vietnamese language, we use the data set from the 2016 VLSP contest. The results show that our model outperforms the best previous systems for Vietnamese NER \cite{nguyen2019} with F1 is 95.61\(\% \) on test set.\\
	The remainder of this paper is structured as follows. Section \ref{secRelated} refers related work on named entity recognition. Section \ref{secMethod} describes the implementation method. Section \ref{secExperiments} gives experimental results and discussions. Finally, the conclusion will be given in Section \ref{secConclusions}.
	\section{RELATED WORK}
	\label{secRelated}
	The approaches for NER task can be divided into two routines: (1) statistical learning approaches and (2) deep learning methods.\\
	In the first type, the authors used traditional labeling models such as crf, hidden markov model, support vector machine, maximum entropy that are heavily dependent on hand-crafted features. Sentences are expressed in the form of a set of features such as word, pos, chunk, etc… Then they are put into a linear model for labeling. Some examples following this routine are \cite{florian2019,lin2009, luo2015,lehong2010}. These models were proven to work quite well for low existing resources languages such as Vietnamese. However, these kinds of NER systems are relied heavily on the used feature set, and on hand-crafted features that are expensive to construct and are difficultly reusable \cite{nguyen2019}.\\
	For the second routine, with the appearance of deep learning models with superior computational performance seems to improve the accuracy of the NER task. The performance of deep learning models also have been shown much better than the statistical based methods. 
	In particular, the convolutional neural network (CNN) \cite{zhang1988}, recurrent neural network (RNN), LSTM networks are popular use, we can exploit the syntax feature through character embedding in combination with word embedding \cite{pham2017-2,wu2019}. Other information such as pos-tag and chunk-tag is also used to provide more information about semantic \cite{pham2017,minh2018-2,anh2019}. The word vectors are combined in different ways, then feed into the Bi-LSTM network with CRF in output. For Vietnamese, there are many NER systems using LSTM network. In \cite{pham2017}, the authors introduced a model that uses two Bi-LSTM layers with softmax layers at the output, with input from vectors using syntax specific, F1 score is 92.05\(\% \). A model using single Bi-LSTM layer combining crf at the output to achieve F1-score of 83.25\% was given in \cite{nguyen2016-2}. A number of high-precision models are introduced in the \cite{anh2019} with Bi-LSTM-CRF model with the input is the extracted vector with characteristic of word character, F1 is 94.88\%. And most recently, a combination of Bi-LSTM - attention layer - CRF model with F1 score of 95.33\% was given in \cite{nguyen2019}.
	\section{METHODOLOGY}
	\label{secMethod}
	\subsection{Feature engineering}
	\label{subsecFeatureE}
	\textbf{Word embedding} To build up a word embedding set, we use the skip-gram neural network model, which is trained from one million articles in 2018 through a Vietnamese news portal (baomoi.com). Skip-gram is used to predict the context word for a given target word. We choose sliding window of Size 2, therefore, normally there are four context words corresponding to a target word. For words that are not trained, a vector called unknown (UNK) embedding is used instead. The UNK embedding is created by random vectors sampled uniformly from the range \([ - \sqrt {\frac{3}{{\dim }}} , + \sqrt {\frac{3}{{\dim }}} ]\), where the dimension (dim) is the dimension of word embeddings \cite{anh2019}. To improve the performance of our system, we use semantic features to vectorize words into models (part of speech tagging and chunk-tag). The pos-tagging vectors and chunk-tag vectors were represented as one-hot vectors.
	\subsubsection{Character Embedding} 
\begin{figure}[htbp]
\label{fig1}
	\begin{center}
		\includegraphics[width=\linewidth]{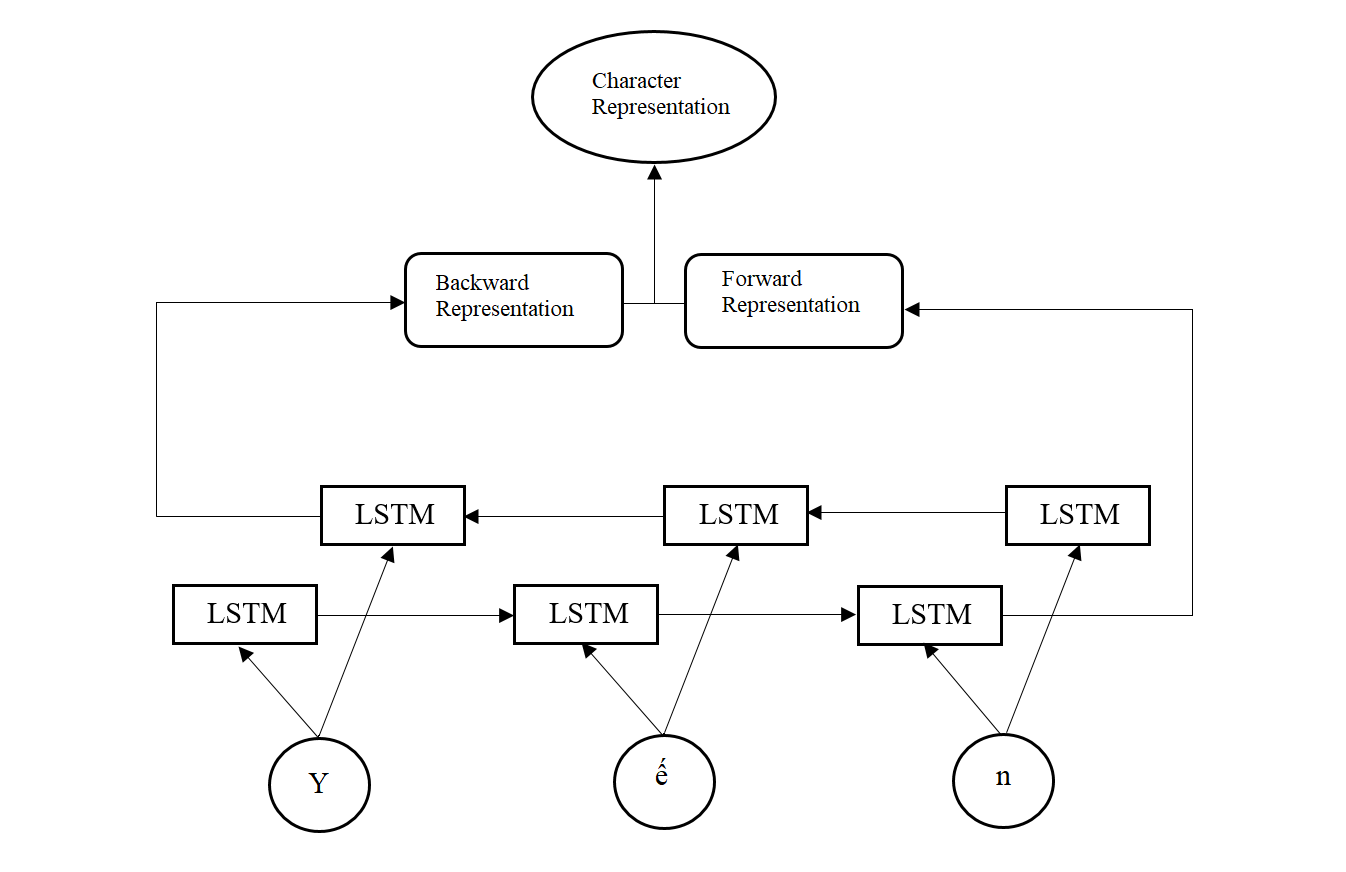}
		\caption{Character-level Embedding}
	\end{center}
\end{figure}

Recently, automatically extracting hidden features using neural networks (LSTM, CNN) used in many articles, proved effective in the NER task \cite{lample2016}. In this research, we use the Bi-LSTM network to extract hidden patterns that characterize the syntax of words, as shown in Fig. \ref{fig1}.
\subsection{LSTM}
LSTM networks are a type of Recurrent Neural Networks (RNN) that uses special units in addition to standard units. LSTM units contain a “memory cell” that can maintain information in memory for controlled periods of time. A cell in the LSTM network consists of three control gates: forget gate (determining which information is ignored and which is retained), update gate (deciding how much of the memorized information is added to the current state) and the output gate (making the decision about which part of the current cell makes it to the output). At time $t$, cell updates are given as follows:
	\begin{align}
	f_t&= \sigma \left( W_f h_{t - 1} + U_f x_t + b_f \right)\\
	i_t&=\sigma (W_i h_{t-1}+U_i x_t+b_i)\\
	o_t&=\sigma (W_0 h_{t-1}+U_0 x_t+b_0)\\
	\mathop c\limits^ \sim  _t&=\tanh(W_c h_{t-1}+U_c x_t+b_c)\\
	c_t&=f_t\odot c_{t-1}+i_t\odot\mathop c\limits^ \sim  _t\\
	h_t&=o_t\odot \tanh(c_t),
	\end{align}
where \(\sigma \) is the sigmoid function and \( \odot \)  is an pointwise operator, which can be multiplication or addition or tanh function, \({x_t}\) is the input vector at time $t$, \({h_t}\) is the hidden state vector that holds information from the beginning to the present time. The gates $f$, $i$, $o$, $c$ are the forget gate, input gate, output gate and cell vector respectively. The matrices \({U_f}\), \({U_i}\),  \({U_o}\),  \({U_c}\) are weight matrices that connect input   and gates. The matrices \({W_f}\), \({W_i}\),  \({W_o}\),  \({W_c}\) are weight matrices that connect gates and hidden state.
	\subsection{Bi-LSTM}
\begin{figure}[htbp]
	\label{fig2}
	\begin{center}
		\includegraphics[width=\linewidth]{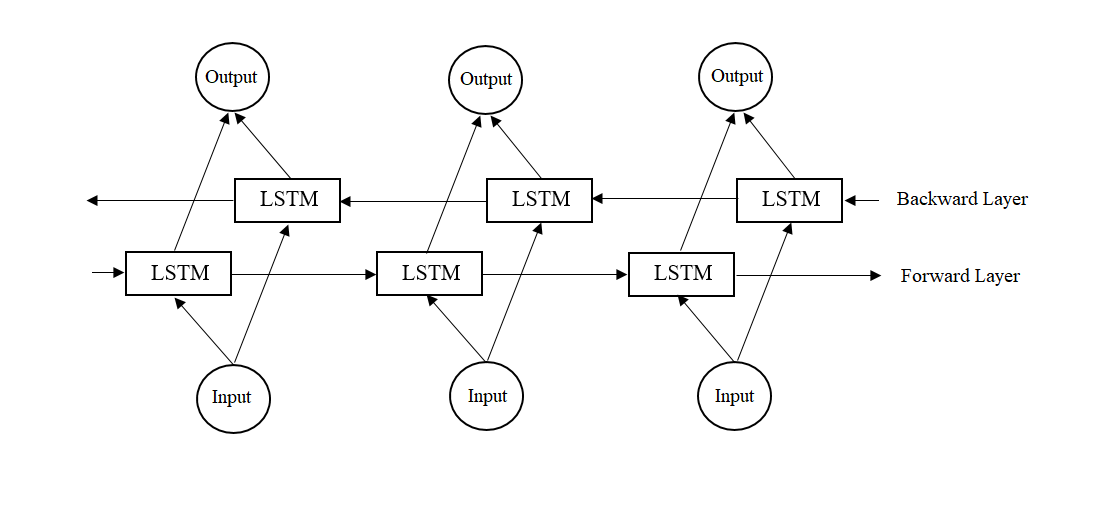}
		\caption{Bidirectional - LSTM}
	\end{center}
	\end{figure}

	In sequence labeling task, the context of a word is represented more effectively by the context, i.e. the companion words, the left and right words in a sentence. To archive these information, a candidate model is Bidirectional-LSTM network as shown in Fig. \ref{fig2}. Then, the output of the Bi-LSTM network is observed by concatenating its left and right context representations.
	
	\subsection{Conditional Random Fields (CRFs)}
	For many sequence labeling tasks, a simple but effective approach is consider the correlation between the labels close together and the best sequence decoding with simple rules. The conditional random field is essentially a probability model, which can predict labels with predefined structures. Instead of decoding independent labels, CRFs learn the sequence of labels from train data to decode output labels at the same time. In CRF, when given a word sequence \(x = ({x_1},{x_2},...,{x_m})\), the conditional probability of a tag sequence \(y = ({y_1},{y_2},...,{y_m})\), is defined as in \cite{minh2018}:
	\begin{align}
	P(y|x) = \frac{{\exp (w.F(y,x))}}{{\sum\nolimits_{y' \in Y} {\exp (w.F(y',x))} }},
	\end{align}
	where $w$ is the parameter vector estimated from training data. The feature function \(\;F(y,x) \in I{R^d}\) is defined globally on an entire input sequence and an entire tag sequence. Space $Y$ is the space of all possible tag sequences. The feature function \(\;F(y,x)\) is calculated by summing local feature functions:
	\begin{align}
	{F_j}(y,x) = \sum\limits_{i = 1}^n {{f_j}({y_{i - 1}},{y_i},x,i)}
	\end{align}
	\subsection{Our Deep Learning Model}
\begin{figure}[htbp]
	\label{fig3}
	\begin{center}
		\includegraphics[width=\linewidth]{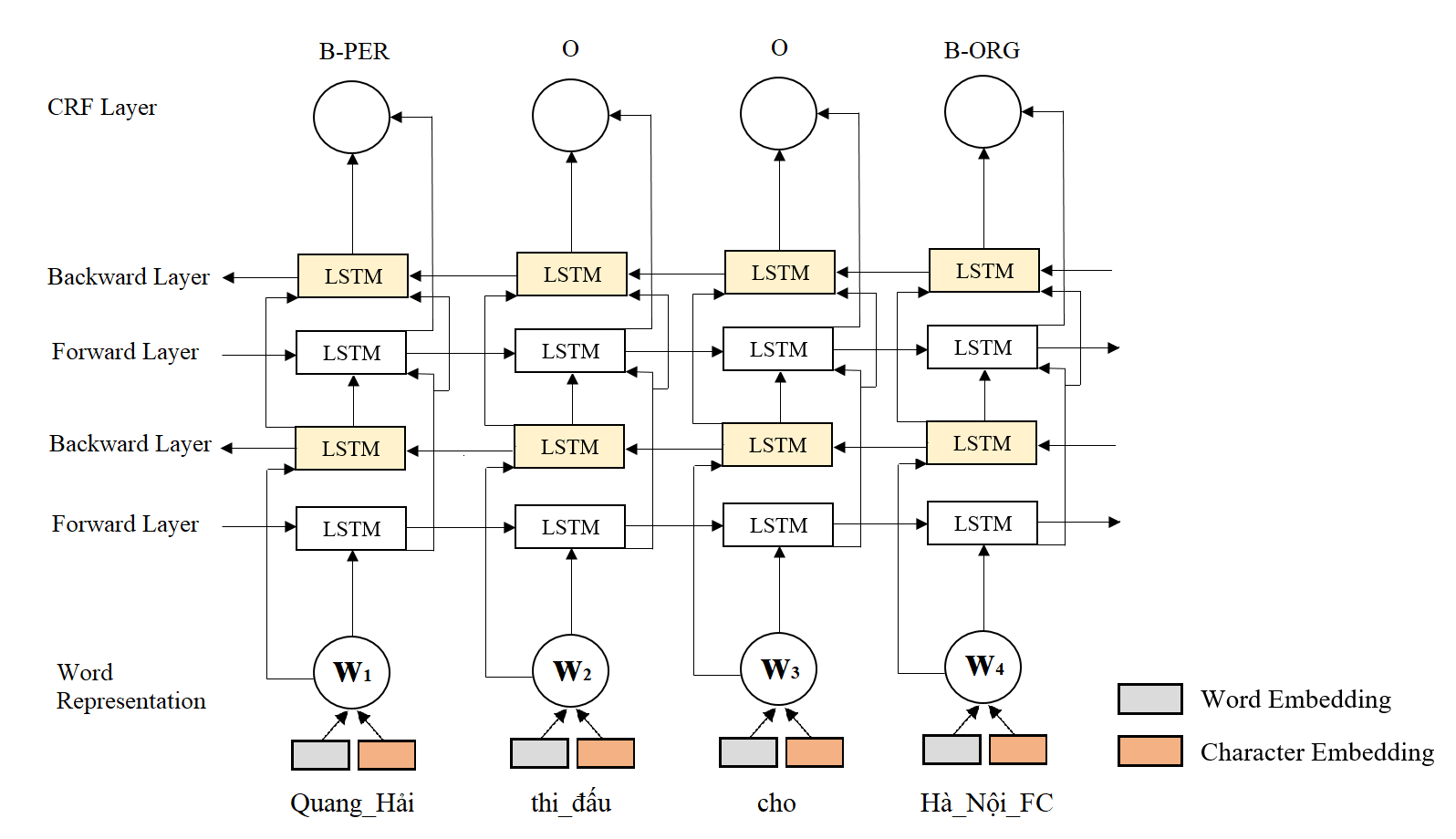}
		\caption{Our Deep Learning model}
	\end{center}
\end{figure}

	For the NER labeling task for Vietnamese, we use multiple Bi-LSTM layers with the CRF layer at the top to detect entities named in the sequence \cite{anh2019}, as shown in Fig. \ref{fig3}. The architecture operates as in the following sequence:
	\begin{itemize}
		\item The input of our neural network is sequence of word representations
		\item Each word representation encoded through two Bi-LSTM layers
		\item The CRF layer at the top to decode hidden feature vectors from previous layer (Bi-LSTM layer)
	\end{itemize}
	
	\section{EXPERIMENTS}
	\label{secExperiments}
	\subsection{Datasets}
	To evaluate the model, we use dataset from VLSP-2016 NER task \cite{vlsp2016}, with four types of entities including person (PER), location (LOC), organization (ORG), miscellaneous (MISC). In addition, VLSP-2016 dataset provides the information about word segmentation, part-of-speech, and chunking tags. The number of sentences in train-set, validation-set and test-set is shown in the T Table \ref{tab1}.
	\begin{table}[htbp]
		\label{tab1}
		\caption{NUMBER OF SENTENCES}
		\begin{center}
			\begin{tabular}{lr}
				\hline 
				\bf Dataset	& \bf Number of sentences\\ 
				\hline 
				Train & 14861\\
				Validation & 2000\\
				Test & 2831\\
				\hline 
				
			\end{tabular}
		
		\end{center}
	\end{table}
	
	The number of entities included in the train set and test set is shown in the following table:
	\begin{table}[htbp]
		\label{tab2}	
		\caption{NUMBER OF LABELS IN DATASET}
		\begin{center}
			\begin{tabular}{lcc}
				\hline 
				\bf Type	& \bf Train & \bf Test\\ 
				\hline 
				Location & 6245 & 1379\\
				Organization & 1213 &	274\\
				Person & 	7480&	1294\\
				Miscellaneous names & 	282&	49\\
				Total & 	15220&2996\\ 
				\hline	
			\end{tabular}
			
		\end{center}
	\end{table}
	
	\subsection{Hyper-parameters}
	Table \ref{tab3} summarizes the hyper-parameters that we have chosen for our NER model. In order to have more efficient training process, the parameters are optimized using Nesterov-accelerated Adaptive Moment Estimation (Nadam) optimizer \cite{dozat2016} with batch size 64. Word representation concatenated by a 300 dimensional word2vec-vector (pre-trained from baomoi.com) and two one-hot vectors represent pos tags and chunks, respectively and a 60 dimensional character vector (generated from a bi-lstm network with dropout rate equal 0.3, as shown in Figure 1). To prevent overfitting, we fix dropout rate to 0.5 for both Bi-LSTM layers (as shown in Figure 3).\\
	The NER model is trained in 40 epoch. First 20 epochs, the initial learning rate is set at 0.004. In the remaining epochs, it is fixed to 0.0004. The best model obtained when the value of the loss function on validation-set is minimal.
	\begin{table}[htbp]
		\label{tab3}
		\caption{THE MODEL HYPE-PARAMETERS}
		\begin{center}
			\begin{tabular}{lr}
				\hline 
				\bf Hyper-parameter	& \bf Value\\ 
				\hline 
				Character dimension &60\\ 
				Word dimension&	300\\
				Hidden size char&30\\
				Hidden size word&64\\
				Update function&Nadam\\
				Learning rate first 20 0epoch&0.004\\
				Learning rate last 20 epoch&0.0004\\
				Dropout character embedding&0.3\\
				Dropout two Bi-LSTM layers&0.5\\
				Batch size&64\\
				\hline
			\end{tabular}
			\label{tab1}
		\end{center}
	\end{table}
	
	\subsection{Experimental Results}
	The experiment is conducted by combining all input features include pos-tag feature, chunk feature and character feature. The results are shown in Table \ref{tab4}.
	
	\begin{table}[htbp]
		\label{tab4}
		\caption{RESULTS ON VLSP 2016 TEST-SET}
		\begin{center}
			\begin{tabular}{lccc}
				\hline 
				&\bf Precision& \bf	Recall& \bf	F1-Score\\ 
				\hline 
				LOC&95.43&	96.95&	96.18\\
				PER&95.53&	97.53&	96.52\\
				ORG&87.32	&90.51&	88.89\\
				MISC&100.0	&87.76&	93.48\\
				Avg/total&95.32	&95.93&	95.61\\
				\hline
			\end{tabular}
			\label{tab1}
		\end{center}
	\end{table}
	With VLSP 2016 dataset, the experiment achieved state-of-the-art performances on Vietnamese NER task with 95.61\(\% \) F1-score. Table \ref{tab5} shows the performance of our deep learning model and several published systems on NER task.
	
	\begin{table}[htbp]
		\label{tab5}
		\caption{PERFORMANCES ON VLSP 2016 DATASET}
		\begin{center}
			\begin{tabular}{lc}
			\hline 
				\bf Model& \bf	F1-Score\\ 
				\hline 
				VNER[12] &95.33\\
				Feature-based CRF [10]&93.93\\
				NNVLP [9]&92.91\\
				Nguyen et al. 2018 [21]&94.88\\
				Our NER model&95.61\\
				
				\hline 	
			\end{tabular}
		\end{center}
	\end{table}
	
	The general difference with other systems in Table 5 is that we trained a new word embedding set by word2vec model, described in Subsection \ref{subsecFeatureE}. Moreover, we use two Bi-LSTM layers in order to encode word representations.
	
	\section{CONCLUSIONS}
	\label{secConclusions}
	In this paper, we presented a neural network model for Vietnamese named entity recognition task, which obtains state-of-the-art performance. Experiments on recognize Vietnamese entity in sequence labeling task showed the effectiveness of training a new word embedding set and using two Bi-LSTM layers in order to extract hidden features from word representations. Our results is outperform the best previous systems for Vietnamese Named entity recognition.
	
	\section*{Acknowledgment}
	The first author also has receive the support from Institute of Mathematics, Vietnam Academy of Science and Technology, Year 2019. 	This work is also supported by iCOMM Media \(\&\) Tech, Jsc. We would like to thank the iCOMM RnD team for supported resources and text data that we used during training and experiments our model.
	
\end{document}